\documentclass{hld2025}

\title[Understanding WSD beyond Transformers]{Universal Dynamics of Warmup Stable Decay: understanding WSD beyond Transformers}

\hldauthor{%
\Name{Annalisa Belloni} \Email{abelloni@student.ethz.ch}\\
\addr {MPI-IS Tübingen, Germany, ETH Zürich, Switzerland and Politecnico di Torino, Italy} {}
\AND
\Name{Lorenzo Noci} \Email{lorenzo.noci@inf.ethz.ch}\\
\addr {ETH Zürich, Switzerland}
\AND
\Name{Antonio Orvieto} \Email{antonio@tue.ellis.eu}\\
\addr {MPI-IS, ELLIS Institute Tübingen, Tübingen AI Center, Germany} {}
}

\usepackage{listings}
\usepackage{xcolor}

\begin{document}

\maketitle

\begin{abstract}%
The Warmup Stable Decay (WSD) learning rate scheduler has recently become popular, largely due to its good performance and flexibility when training large language models. It remains an open question whether the remarkable performance of WSD -- using a decaying learning rate for only a fraction of training compared to cosine decay -- is a phenomenon specific to transformer-based language models that can potentially offer new theoretical insights into their training dynamics. Inspired by the usage of learning rate schedulers as a new lens into understanding landscape geometry~(e.g., river valley, connected minima, progressive sharpening), in this work we compare the WSD path of the Adam optimizer on a Pythia-like language model to that of a small CNN trained to classify CIFAR10 images. We observe most training signals, optimizer path features, and sharpness dynamics to be qualitatively similar in such architectures. This consistency points to shared geometric characteristics of the loss landscapes of old and new nonconvex problems, and hints to future research questions around the geometry of high dimensional optimization problems.
\end{abstract}

\section{Introduction}
The growing adoption of the WSD scheduler is driven by its good~(at times, better) performance in relation to standard cosine decay practice, in combination to its native support for resuming training at different scales. The latter, in particular, has become an increasingly urgent concern given the rise of large language models (LLMs), which are extremely costly to train.

To better appreciate the benefits of the WSD scheduler for training continuation, let us briefly revisit the behavior of the well-established warmup cosine scheduler alongside the newer WSD scheduler. Here, \(T_{end}\) denotes the total step budget, \(T_w\) the number of warmup steps, and \(T_c\) the step at which the learning rate decay begins.
The warmup cosine annealing scheduler reads:
\[
\eta(t) =
\begin{cases}
\frac{t}{T_w} & \text{if } t \leq T_w, \\
\frac{1}{2} (1 + \cos (\pi \frac{t - T_w}{T_{end} - T_w} ) ) & \text{if } t > T_w
\end{cases}
\]
The WSD scheduler, instead, has the following form:
\[
\eta(t) =
\begin{cases}
\frac{t}{T_w} & \text{if } t \leq T_w, \\
1 & \text{if } T_w < t \leq T_c, \\
1-\frac{t - T_c}{T_{end} - T_c} & \text{if } t > T_c
\end{cases}
\]
where we assumed a peak learning rate of $1$. Unlike cosine annealing, which requires prior knowledge of the total training duration (i.e., the endpoint \(T_{end}\) must be determined by the time the warmup phase \(T_w\) concludes), the WSD scheduler offers significantly greater flexibility. Specifically, \(T_{end}\) must only be defined at the start of the decay phase \(T_c\), which itself can be selected arbitrarily. This enables a more dynamic training process that can be adapted in real time.

Indeed, as noted by \citet{hu2024minicpmunveilingpotentialsmall} and by \citet{hägele2024scalinglawscomputeoptimaltraining}, the WSD scheduler proves particularly advantageous in the context of \textit{continual training}. It allows training to be resumed from a checkpoint located just prior to the decay phase, retaining the previous high and stable learning rate. This approach eliminates the need to potentially retrain the model from scratch when increasing the training step budget, as would be necessary in a single-cycle cosine annealing schedule; it also avoids the negative effects of rewarming the learning rate, such as loss spikes and unlearning behaviors \cite{ibrahim2024simplescalablestrategiescontinually} \cite{singh2025cosinedecayeffectivenessinfinite}, which may arise when attempting to resume training with a new cycle of cosine annealing.
Given that the primary benefits of WSD lie in compute advantages, it is unsurprising that prior research has focused predominantly on studying its behavior when training large-scale models, particularly transformer-based language models. This narrow focus has shaped a perspective on the scheduler that is deeply linked to the characteristics of transformer-based architectures, in fact some of these studies have even proposed new theoretical insights into the structure of the transformer loss landscapes stemming from WSD dynamics~\cite{wen2024understandingwarmupstabledecaylearningrates}.

\begin{figure}[t]
    \centering
    \vspace{-11mm}
    \includegraphics[width=0.45\textwidth]{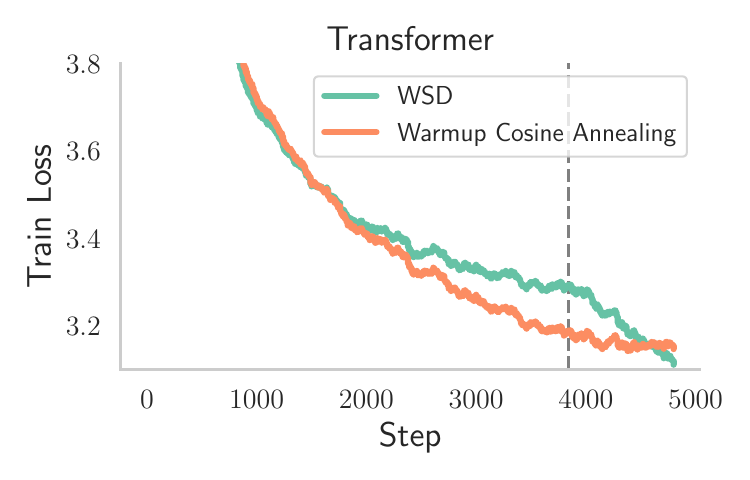}
    \hspace{0.4cm}
    \includegraphics[width=0.45\textwidth]{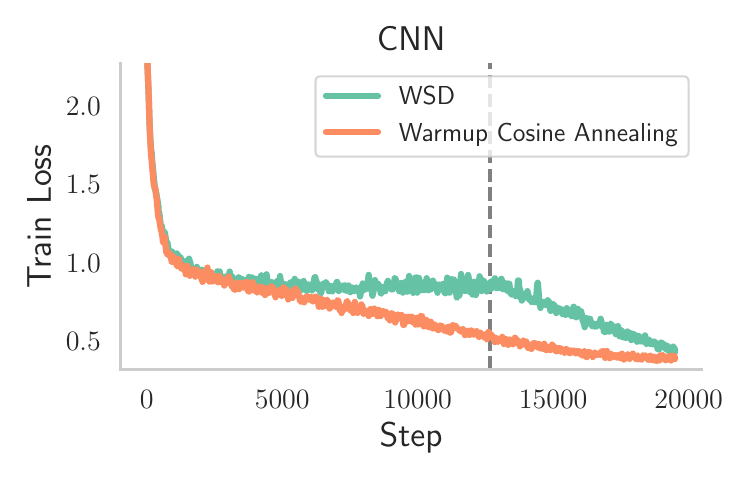}
    \vspace{-5mm}
    \caption{Typical training dynamics with WSD and Warmup Cosine Annealing schedulers. In both the 160M parameters LM with Pythia 12 layer configuration~\citep{biderman2023pythia} (left) and the 334K parameters CNN with 4 layers (right), we observe that WSD follows the same trend: modest decreases during the stable phase (left of dashed line), followed by sudden and more pronounced gains during the cooldown (right of dashed line). Note that while the former model was trained 3B tokens~(way less than 1 epoch), the latter performs 50 epochs.}
    \vspace{-5mm}
    \label{fig:Figure1}
\end{figure}

In contrast, our work investigates the effects of WSD on alternative architectures like CNNs, where only brief references to the \textit{trapezoid} scheduler (an earlier term for WSD) have been made -- for example by \citet{xing2018walksgd} -- none driven by computational efficiency, which has rarely posed a significant constraint on small CNN models.
Interestingly, we observed that WSD induces similar training dynamics even in these different settings (Fig. \ref{fig:Figure1}).
Motivated by this similarity, we aim to investigate whether WSD behavior remains consistent between the two model types even at a more fine-grained scale. Our analysis reveals the presence of broad and shared properties that govern the characteristic loss pattern induced by WSD across a variety of architectures.

\section{Related Work}
In Fig. \ref{fig:Figure1} we recall the typical WSD loss curve characterized by slow improvements during the stable learning rate phase, followed by a sharp decline during the learning rate cooldown phase \cite{hu2024minicpmunveilingpotentialsmall} \cite{hägele2024scalinglawscomputeoptimaltraining} \cite{schaipp2025surprisingagreementconvexoptimization} 
\cite{wen2024understandingwarmupstabledecaylearningrates}
\cite{xing2018walksgd}
\cite{zhai2022scalingvisiontransformers} . WSD can achieve performance comparable to that of the widely used warmup cosine annealing scheduler, and even outperform it when the cooldown phase is sufficiently long \cite{hu2024minicpmunveilingpotentialsmall} \cite{hägele2024scalinglawscomputeoptimaltraining} \cite{wen2024understandingwarmupstabledecaylearningrates}.
This distinctive and non-traditional training loss curve produced by WSD has recently sparked curiosity and interest, prompting several interpretations of this phenomenon.

\citet{wen2024understandingwarmupstabledecaylearningrates} for example attribute the typical loss trend to the geometric properties of the loss landscapes in transformer-based language models. Proposing and assuming that the loss surface exhibits a \textbf{``River Valley''} profile, they show how variations in the learning rate guide the training iterates along a particular trajectory that navigates this valley.
Specifically, they identify two predominant directions of movement, corresponding to the two phases of training: constant learning rate and cooldown. During the constant phase, due to the high learning rate and the stochasticity of the optimizer, the iterates oscillate between the valley's slopes, making some progress in the river direction. In the cooldown phase, instead, as the learning rate decays, the iterates begin descending on the mountain slope and towards the valley floor, revealing most of the loss reduction.
To support the River Valley hypothesis, \citet{wen2024understandingwarmupstabledecaylearningrates} explore why such a landscape might naturally emerge in language models. They argue that it is \textit{rooted in the intrinsic nature of language modeling itself}, particularly due to the ``heterogeneity in the stochasticity of tokens''. Using a simple bigram toy model, they illustrate how more predictable tokens, typically encoding structured knowledge, shape the flat river-like direction in the landscape. In contrast, tokens associated with higher uncertainty, reflecting linguistic ambiguity or creative variability, contribute to the steeper sides of the valley. 

\citet{schaipp2025surprisingagreementconvexoptimization} instead suggest an alternative explanation for the loss behavior with WSD, grounded in first-order \textbf{convex optimization theory}. Specifically, the authors observe a striking similarity between the loss curve of an LLM trained with AdamW and a suboptimality bound for SGD in convex settings (\cite{schaipp2025surprisingagreementconvexoptimization} Fig. 1). Notably, this explanation does not rely on intrinsic properties of the model, even though it was formulated by analyzing the WSD training dynamics on an LLM.

\section{Experiments}
Building on the macroscopic similarity observed in how WSD affects performance in both transformer language models and CNNs for image classification (Fig. \ref{fig:Figure1}), we aim to examine at a finer scale the training dynamics and the loss landscape regions traversed by AdamW iterates under WSD across the two model types.
\paragraph{Setup}
For our empirical evaluation on transformer-based LMs, we train a decoder-only transformer model \cite{vaswani2017attention}, with around 160M trainable parameter, on the cerebras/SlimPajama-627B HF dataset~\citep{cerebras2023slimpajama}, using PlainLM by \citet{ajroldi2024plainlm}.
Extending the study of WSD to non-transformer-based architectures instead, we experiment with a small CNN, with approximately 334K parameters, trained for an image classification task on the CIFAR10 dataset. All models are trained with Adam~\cite{kingma2014adam}. For further details on the experimental setup, please refer to Appendix \ref{app:exp_setup}. 

\paragraph{River Valley}
First, we aim to replicate some of the findings of \citet{wen2024understandingwarmupstabledecaylearningrates}. To support their River Valley hypothesis, \citet{wen2024understandingwarmupstabledecaylearningrates} provide visualizations of the loss landscape by plotting the loss along linear interpolations between selected training checkpoints near the end of the stable phase, obtaining a convex loss curve resembling a valley (\cite{wen2024understandingwarmupstabledecaylearningrates} Fig. 7a). In contrast, when plotting the loss along the interpolation between two points before and after a cooldown (\cite{wen2024understandingwarmupstabledecaylearningrates} Fig. 7b), they observe a smooth decline, consistent with the observations of \citet{hägele2024scalinglawscomputeoptimaltraining} (Fig. 7), who characterize this phase as a smooth transition to a connected basin in the loss landscape.

\begin{figure}
    \centering
    \vspace{-3mm}
        \includegraphics[width=0.48\textwidth]{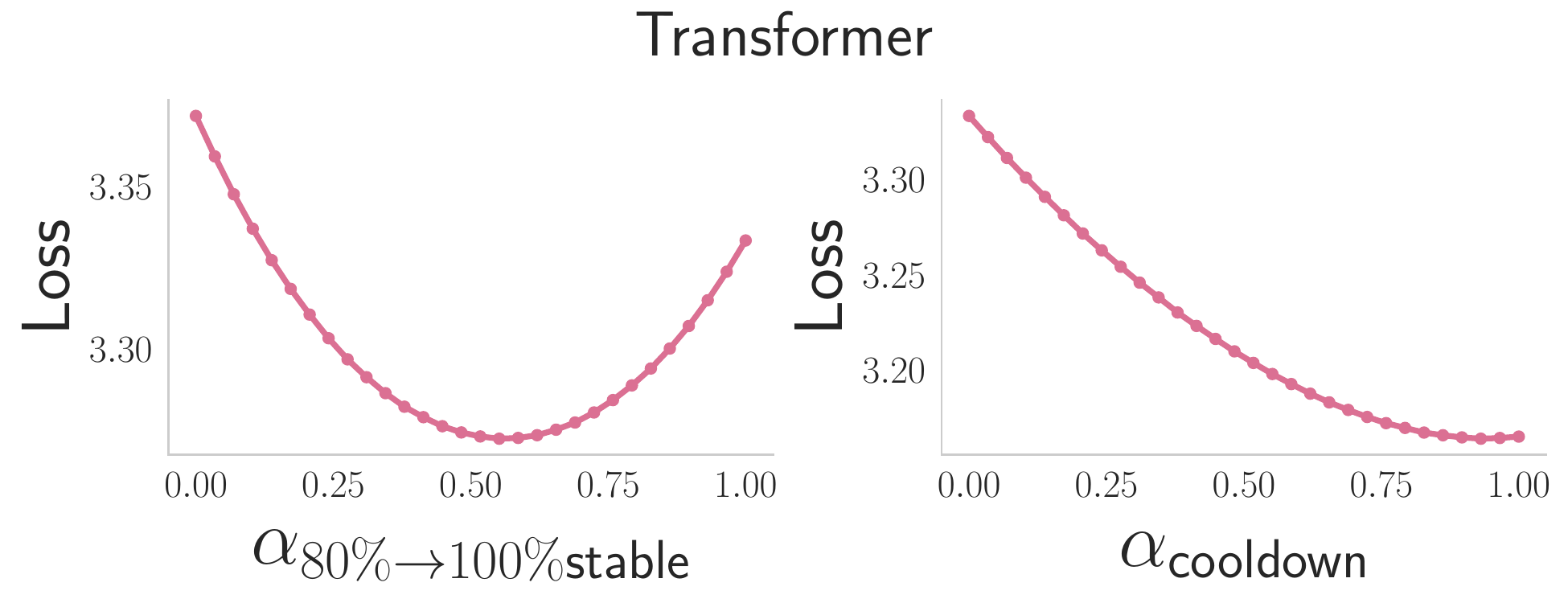}
        \hspace{0.4cm}
        \includegraphics[width=0.48\textwidth]{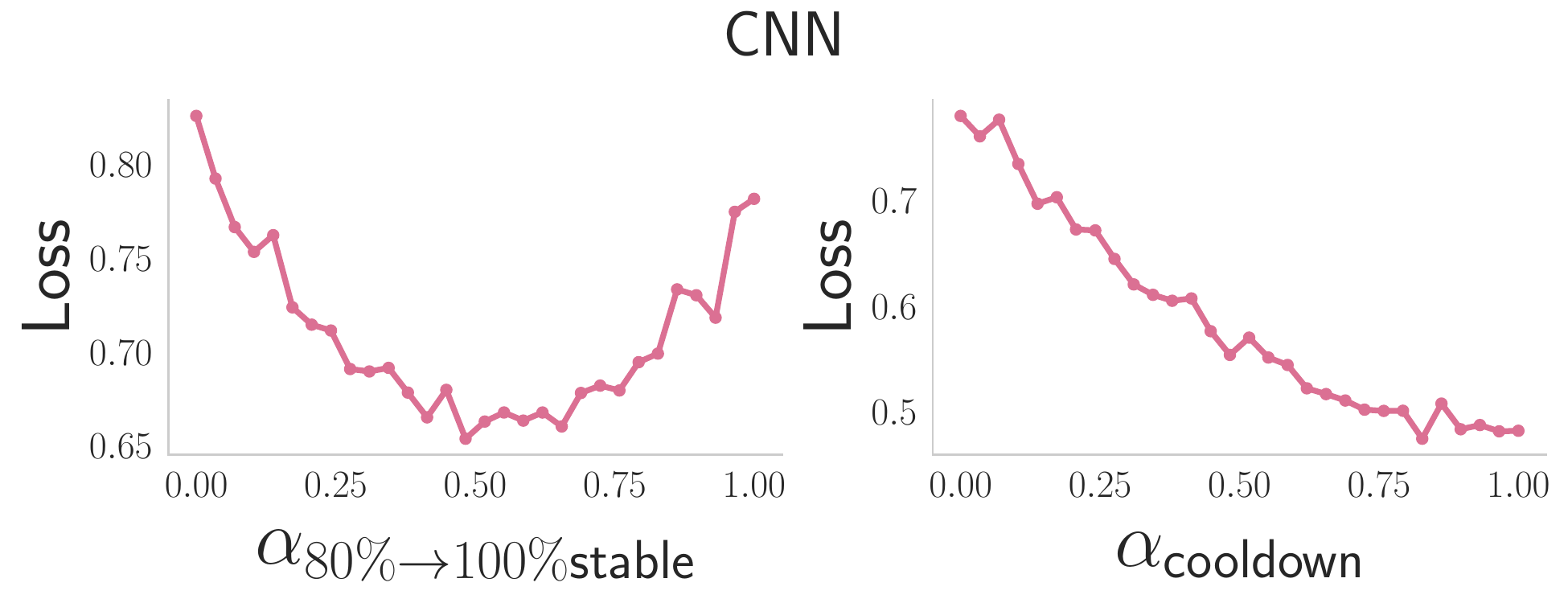}
    \vspace{-9mm}
        \caption{Loss evaluated along $\alpha t + (1 - \alpha)t'$, with $\alpha \in [0,1]$, where $t'$ and $t$ denote checkpoints taken at 80\% and 100\% of the stable phase, or at the start and end of the cooldown phase, on an LM (left) and a CNN (right). Both models display a similar convex valley-shaped profile between the two checkpoints sampled near the end of the stable phase, followed by a monotonic descent over the cooldown period.}
    \vspace{-8mm}
    \label{fig:Figure2}
\end{figure}

We reproduce the same visualizations in Fig. \ref{fig:Figure2}, by interpolating between the checkpoint at 80\% of the stable phase and the one at the start of the cooldown, as well as between the one at the start and at the end of the cooldown itself.
We obtain consistent results not only for transformer but, perhaps surprisingly, also for the CNN.
It therefore appears that the ``river valley'' landscape induced by WSD is not unique to transformer-based language models. Indeed, the loss surface of CNNs also seems to exhibit this profile within the optimizer’s region under WSD, which may partially align with some findings of \citet{xing2018walksgd} (although obtained using SGD and GD, as opposed to Adam).

\begin{figure}[t]
\centering
\vspace{-3mm}
\includegraphics[height=0.17\linewidth]{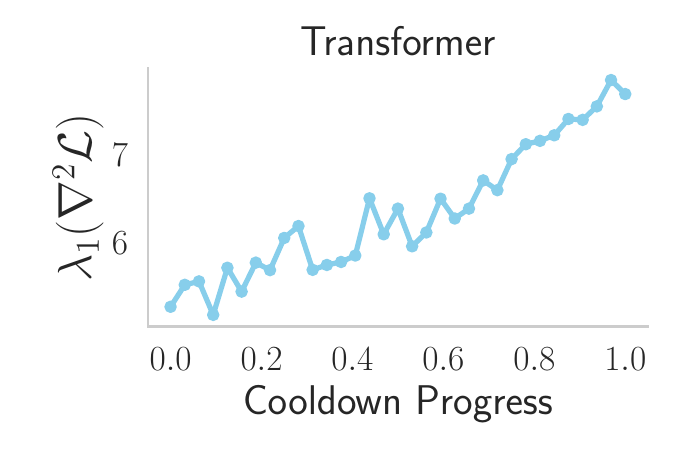}
\includegraphics[height=0.17\linewidth]{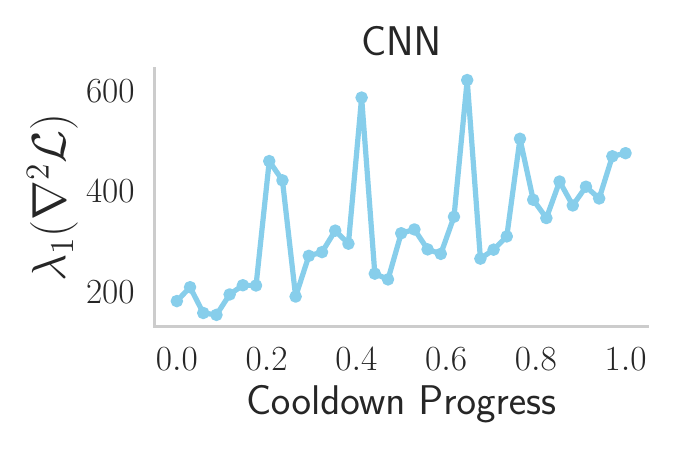}
\includegraphics[height=0.17\linewidth]{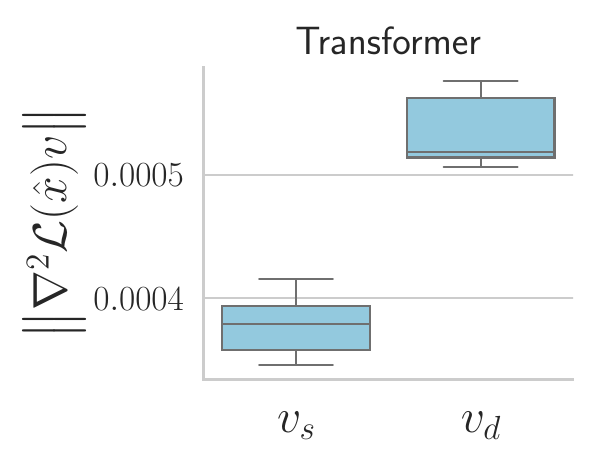}
\includegraphics[height=0.17\linewidth]{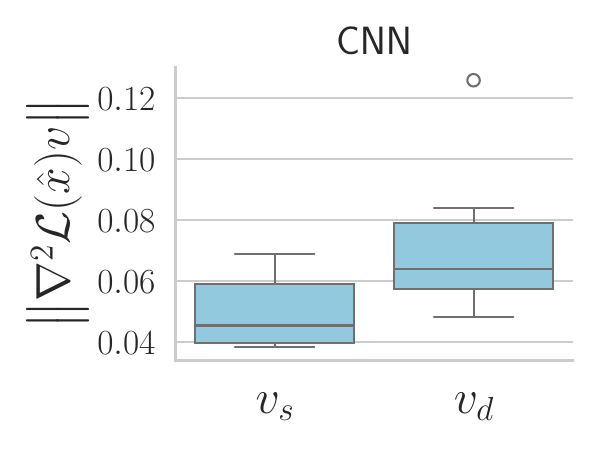}
\vspace{-4mm}
\caption{\textbf{(Left)} Estimated sharpness, evaluated on iterates \(x_i\) sampled at a fixed rate along the cooldown phase, on an LM (left) and on a CNN (right). In both models, reducing the learning rate leads to sharper regions of the loss surface. \textbf{(Right)} Box plots of \(\left\| \nabla^2 \mathcal{L}(\hat{x}) v \right\|\) for directions $v_s$ and $v_d$, derived from multiple training runs for an LM (left) and a CNN (right), where \(\hat{x}\) is the checkpoint at the start of the decay. Results indicate that the direction representing the early decay phase $v_d$ aligns more with high-curvature regions.}
\label{fig:Figure3}
\end{figure}

\paragraph{Sharpness} Further examining the shape of the loss function, we observe that, for both models, sharpness (the largest eigenvalue of the loss Hessian) tends to increase along the iterates collected while annealing the learning rate~(which we name $D$), consistent with the findings of \citet{hu2024minicpmunveilingpotentialsmall}, and more in general with \citet{cohen2022gradientdescentneuralnetworks} and \citet{jastrzebski2019relationsharpestdirectionsdnn}, as shown in Fig. \ref{fig:Figure3}.

\paragraph{Training Directions}
Focusing on the training trajectory, we aim to verify if the stable and cooldown phases correspond to two distinct movement directions in the parameter space, as suggested by \citet{wen2024understandingwarmupstabledecaylearningrates}. To investigate this, we perform Principal Component Analysis (PCA) on two sets, \(S\) and \(D\), each containing iterates \(x_i\) uniformly sampled during stable and decay phases, respectively.
For both models and phases, the first component captures at least approximately 40\% of the total variance (Fig. \ref{fig:Figure7}), suggesting that each phase is primarily governed by a main yet distinct direction.

\paragraph{Cooldown unveils a sharp tunnel}
Motivated by this common trait and by the sharpness increase at cooldown, we zoom in on checkpoint \(\hat{x}\) at decay start, to closely examine how and why sharpness evolves along the optimizer's path.
Given two sets, \(S'\) and \(D'\), consisting of iterates \(x_i\) uniformly sampled from the last 20\% of the stable and the first 20\% of the cooldown, respectively, we center both point clouds and apply PCA to each. This yields two main parameter-space directions, \(v_s\) and \(v_d\), which capture the trajectory near the loss curve elbow.
We then analyze how these directions align with the Hessian \(\nabla^2 \mathcal{L}(\hat{x})\) eigenspaces, finding
\(
\left\| \nabla^2 \mathcal{L}(\hat{x}) v_s \right\| < \left\| \nabla^2 \mathcal{L}(\hat{x}) v_d \right\|
\) for both models (Fig. \ref{fig:Figure3}).
This suggests that the direction corresponding to the early cooldown lies more closely in high-curvature regions of the loss landscape.
During decay, despite much smaller updates (Fig. \ref{fig:Figure6}) \cite{hu2024minicpmunveilingpotentialsmall}, true loss minimization becomes clear as the reduced step size allows the trajectory to ``see'', and thus to follow, sharper subspaces of the parameter space, that were previously not accessible.

\begin{figure}
\vspace{-3mm}
    \centering
    \includegraphics[width=0.23\textwidth]
    {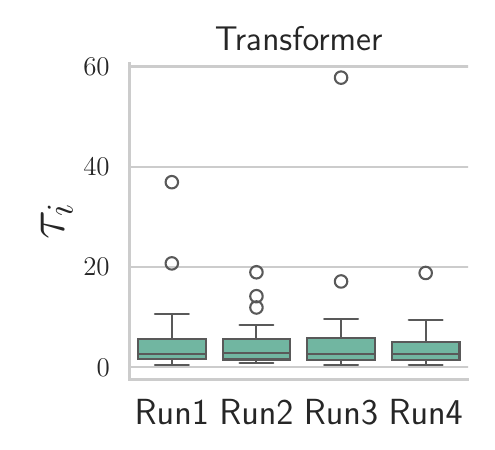}
    \includegraphics[width=0.23\textwidth]{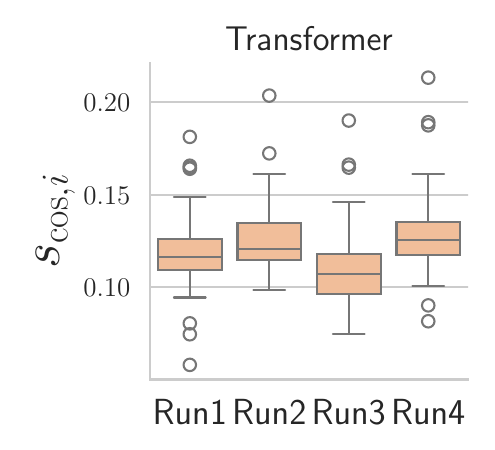}
    \hspace{0.4cm}
    \includegraphics[width=0.23\textwidth]{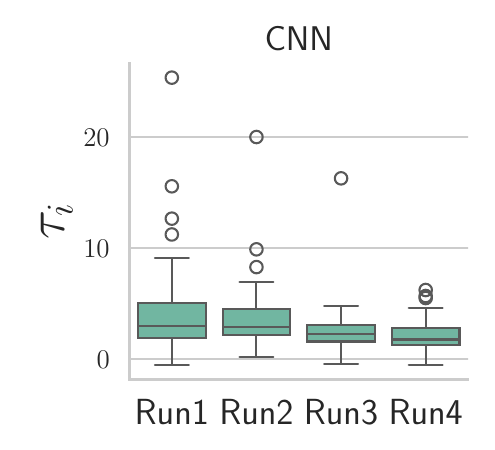}
    \includegraphics[width=0.23\textwidth]{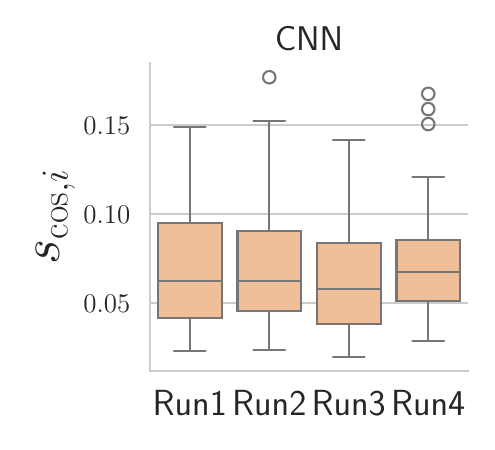}
    \vspace{-3mm}
    \caption{\textbf{(Green)} Box plot of \(\tau_i = \frac{-\nabla \mathcal{L}(x_i)(x^* - x_i)}{\mathcal{L}(x_i) - \mathcal{L}(x^*)}\), evaluated on a a subset \(B\) of iterates \(x_i\) sampled regularly during training, with \(x^*\) denoting the solution at the end of training. Under weak quasi-convexity, \(\tau_i > 0\), which is indeed observed in nearly all cases.
    \textbf{(Orange)} Box plot of \(s_{\cos,i}\) = \(\text{cos\_similarity}(-\nabla \mathcal{L}(x_i),\, x_{i+1} - x_i)\), evaluated on \(B\). These scores remain positive, suggesting that AdamW updates do not deviate substantially from those of SGD.
    Results are presented for multiple training runs for an LM (left) and a CNN (right).} 
    \vspace{-4mm}
    \label{fig:Figure4}
\end{figure}

\paragraph{Quasi-Convexity}
Finally, we end by examining the broader explanation from \citet{schaipp2025surprisingagreementconvexoptimization}, which currently shows a mismatch between the application context (non-convex loss optimized with AdamW) and the theoretical setting on which the argument is based (convex loss optimized with SGD).
Therefore, we investigate how ``surprising'' the alignment between the loss profile and the theoretical bound truly is.
We test the \textit{Weak-Quasi-Convexity} condition introduced by \citet{hardt2019gradientdescentlearnslinear} (Def. 2.1), evaluating it over a domain \(B\) consisting of a subset of iterates \(x_i\) sampled regularly after warmup. As shown in Fig. \ref{fig:Figure4}, \textit{Weak-Quasi-Convexity} holds for both models, in nearly all cases.
Similarly, we compare AdamW updates to those of SGD. To do so, we compute the cosine similarity between the negative gradient and the actual update vector at each point in \(B\), consistently finding positive values for both models, as presented in Fig. \ref{fig:Figure4}. 
Taken together, these results reduce the perceived surprise regarding the alignment observed by \citet{schaipp2025surprisingagreementconvexoptimization}, as the loss appears to exhibit a nearly convex behavior along AdamW's trajectory. So, these last findings provide validation for an explanation of WSD that seems to hold regardless of the architecture, and they also point out other potentially interesting analogies between transformer LMs and CNNs.

\section{Discussion and Conclusions}
The experiments and results obtained suggest that the distinctive behavior of WSD, although primarily observed and studied in transformer-based language models, is not exclusive to them. This is evident not only in overall performance, but also through a closer analysis of the training dynamics and of the loss surface structure: we observe general and shared characteristics that support the typical WSD loss curve across two distinct~(in size and training modality) architectures. An intriguing future direction is to investigate the effect of overparametrization on WSD performance, not indeed that our CNN, being small, does not perfectly fit the data~(similarly to the transformer).

\section*{Acknowledgments}
The authors thank the Hector Foundation for the financial support provided.

\newpage
\bibliography{main}

@misc{cerebras2023slimpajama,
author = {Soboleva, Daria and Al-Khateeb, Faisal and Myers, Robert and Steeves, Jacob R and Hestness, Joel and Dey, Nolan},
title = {{SlimPajama: A 627B token cleaned and deduplicated version of RedPajama}},
month = June,
year = 2023,
howpublished = {\url{https://cerebras.ai/blog/slimpajama-a-627b-token-cleaned-and-deduplicated-version-of-redpajama}},
url = {https://huggingface.co/datasets/cerebras/SlimPajama-627B},
}

@article{kingma2014adam,
  title={Adam: A method for stochastic optimization},
  author={Kingma, Diederik P},
  journal={arXiv preprint arXiv:1412.6980},
  year={2014}
}

@inproceedings{biderman2023pythia,
  title={Pythia: A suite for analyzing large language models across training and scaling},
  author={Biderman, Stella and Schoelkopf, Hailey and Anthony, Quentin Gregory and Bradley, Herbie and O’Brien, Kyle and Hallahan, Eric and Khan, Mohammad Aflah and Purohit, Shivanshu and Prashanth, USVSN Sai and Raff, Edward and others},
  booktitle={International Conference on Machine Learning},
  pages={2397--2430},
  year={2023},
  organization={PMLR}
}

@misc{hu2024minicpmunveilingpotentialsmall,
      title={MiniCPM: Unveiling the Potential of Small Language Models with Scalable Training Strategies}, 
      author={Shengding Hu and Yuge Tu and Xu Han and Chaoqun He and Ganqu Cui and Xiang Long and Zhi Zheng and Yewei Fang and Yuxiang Huang and Weilin Zhao and Xinrong Zhang and Zheng Leng Thai and Kaihuo Zhang and Chongyi Wang and Yuan Yao and Chenyang Zhao and Jie Zhou and Jie Cai and Zhongwu Zhai and Ning Ding and Chao Jia and Guoyang Zeng and Dahai Li and Zhiyuan Liu and Maosong Sun},
      year={2024},
      eprint={2404.06395},
      archivePrefix={arXiv},
      primaryClass={cs.CL}
}

@misc{hägele2024scalinglawscomputeoptimaltraining,
      title={Scaling Laws and Compute-Optimal Training Beyond Fixed Training Durations}, 
      author={Alexander Hägele and Elie Bakouch and Atli Kosson and Loubna Ben Allal and Leandro Von Werra and Martin Jaggi},
      year={2024},
      eprint={2405.18392},
      archivePrefix={arXiv},
      primaryClass={cs.LG}
}

@misc{schaipp2025surprisingagreementconvexoptimization,
      title={The Surprising Agreement Between Convex Optimization Theory and Learning-Rate Scheduling for Large Model Training}, 
      author={Fabian Schaipp and Alexander Hägele and Adrien Taylor and Umut Simsekli and Francis Bach},
      year={2025},
      eprint={2501.18965},
      archivePrefix={arXiv},
      primaryClass={cs.LG}
}

@misc{wen2024understandingwarmupstabledecaylearningrates,
      title={Understanding Warmup-Stable-Decay Learning Rates: A River Valley Loss Landscape Perspective}, 
      author={Kaiyue Wen and Zhiyuan Li and Jason Wang and David Hall and Percy Liang and Tengyu Ma},
      year={2024},
      eprint={2410.05192},
      archivePrefix={arXiv},
      primaryClass={cs.LG}
}

@misc{ibrahim2024simplescalablestrategiescontinually,
      title={Simple and Scalable Strategies to Continually Pre-train Large Language Models}, 
      author={Adam Ibrahim and Benjamin Thérien and Kshitij Gupta and Mats L. Richter and Quentin Anthony and Timothée Lesort and Eugene Belilovsky and Irina Rish},
      year={2024},
      eprint={2403.08763},
      archivePrefix={arXiv},
      primaryClass={cs.LG}
}

@misc{zhai2022scalingvisiontransformers,
      title={Scaling Vision Transformers}, 
      author={Xiaohua Zhai and Alexander Kolesnikov and Neil Houlsby and Lucas Beyer},
      year={2022},
      eprint={2106.04560},
      archivePrefix={arXiv},
      primaryClass={cs.CV}
}

@misc{ajroldi2024plainlm,
  author = {Niccolò Ajroldi},
  title = {plainLM: Language Model Pretraining in PyTorch},
  year = {2024},
  howpublished = {\url{https://github.com/Niccolo-Ajroldi/plainLM}}
}

@misc{hardt2019gradientdescentlearnslinear,
      title={Gradient Descent Learns Linear Dynamical Systems}, 
      author={Moritz Hardt and Tengyu Ma and Benjamin Recht},
      year={2019},
      eprint={1609.05191},
      archivePrefix={arXiv},
      primaryClass={cs.LG}
}

@misc{cohen2022gradientdescentneuralnetworks,
      title={Gradient Descent on Neural Networks Typically Occurs at the Edge of Stability}, 
      author={Jeremy M. Cohen and Simran Kaur and Yuanzhi Li and J. Zico Kolter and Ameet Talwalkar},
      year={2022},
      eprint={2103.00065},
      archivePrefix={arXiv},
      primaryClass={cs.LG}
}

@misc{jastrzebski2019relationsharpestdirectionsdnn,
      title={On the Relation Between the Sharpest Directions of DNN Loss and the SGD Step Length}, 
      author={Stanislaw Jastrzebski and Zachary Kenton and Nicolas Ballas and Asja Fischer and Yoshua Bengio and Amos Storkey},
      year={2019},
      eprint={1807.05031},
      archivePrefix={arXiv},
      primaryClass={stat.ML}
}

@misc{xing2018walksgd,
      title={A Walk with SGD}, 
      author={Chen Xing and Devansh Arpit and Christos Tsirigotis and Yoshua Bengio},
      year={2018},
      eprint={1802.08770},
      archivePrefix={arXiv},
      primaryClass={stat.ML}
}

@article{vaswani2017attention,
  title={Attention is all you need},
  author={Vaswani, Ashish and Shazeer, Noam and Parmar, Niki and Uszkoreit, Jakob and Jones, Llion and Gomez, Aidan N and Kaiser, {\L}ukasz and Polosukhin, Illia},
  journal={Advances in neural information processing systems},
  volume={30},
  year={2017}
}

@misc{singh2025cosinedecayeffectivenessinfinite,
      title={Beyond Cosine Decay: On the effectiveness of Infinite Learning Rate Schedule for Continual Pre-training}, 
      author={Vaibhav Singh and Paul Janson and Paria Mehrbod and Adam Ibrahim and Irina Rish and Eugene Belilovsky and Benjamin Thérien},
      year={2025},
      eprint={2503.02844},
      archivePrefix={arXiv},
      primaryClass={cs.LG}, 
}

\newpage
\appendix

\section{Experimental Setup}
\label{app:exp_setup}
In our empirical analysis of language models, we use a transformer-based language model \cite{vaswani2017attention}, with just over 160M trainable parameters.  We train it on approximately 3 billion tokens from the the cerebras/SlimPajama-627B HF dataset ~\citep{cerebras2023slimpajama}, employing a batch size of 256 and a sequence length of 2048, using PlainLM \cite{ajroldi2024plainlm}.\\
To examine WSD training dynamics in non-transformer architectures instead, we use as a case study a small CNN with about 334K parameters. We train it for an image classification task on the CIFAR10 dataset \url{https://www.cs.toronto.edu/~kriz/cifar.html}, using a batch size of 128. For completeness, we provide below its architecture implemented in PyTorch.
\begin{lstlisting}[label={lst:CNN_archit}]
import torch.nn as nn
import torch.nn.functional as F

class CIFARCNN2(nn.Module):
    def __init__(self):
        super().__init__()
        self.pool = nn.MaxPool2d(2, 2)
        self.conv1 = nn.Conv2d(3, 32, 3, padding = 1)
        self.conv2 = nn.Conv2d(32, 128, 3, padding = 1)
        self.conv3 = nn.Conv2d(128, 128, 3, padding = 1)
        self.conv4 = nn.Conv2d(128, 128, 3, padding = 1)
        self.fc1 = nn.Linear(128, 10)

    def forward(self, x):
        x = self.pool(F.relu(self.conv1(x)))
        x = self.pool(F.relu(self.conv2(x)))
        x = self.pool(F.relu(self.conv3(x)))
        x = self.pool(F.relu(self.conv4(x))) 
        x = x.mean(dim=[2,3]) 
        x = self.fc1(x)
        return x

\end{lstlisting}

To enable meaningful comparisons between such fundamentally different models, it is important to keep the following caveat in mind. CNNs, by the end of training, typically operate in a near-zero loss region of the landscape, having been exposed to the same training data over multiple epochs. In contrast, language models remain far from zero loss at that point as they rarely, if ever, revisit the same sequences. This crucial difference results in distinct training dynamics and loss surface behaviors.
For this reason, we intentionally avoid training the CNN to full accuracy, specifically training it for about 50 epochs. This choice allows for more consistent comparisons, avoiding situations where the two models reach entirely different regions of their respective loss landscapes. It also helps prevent overfitting and limits the influence of secondary effects that tend to emerge when a model reaches perfect accuracy.\\
Both models are also trained using the AdamW optimizer.
Finally, we mention that in our experiments on LM, we follow \citet{schaipp2025surprisingagreementconvexoptimization} and use a cooldown length of 20\% of the total steps. On CNN instead, a longer decay (35\%) is needed to better match warmup cosine annealing performance. Additionally, both WSD and warmup cosine annealing schedulers decay the learning rate to zero for consistency.

\section{Supplementary Figures}

\begin{figure}[h]
    \centering
    \includegraphics[width=0.32\textwidth]{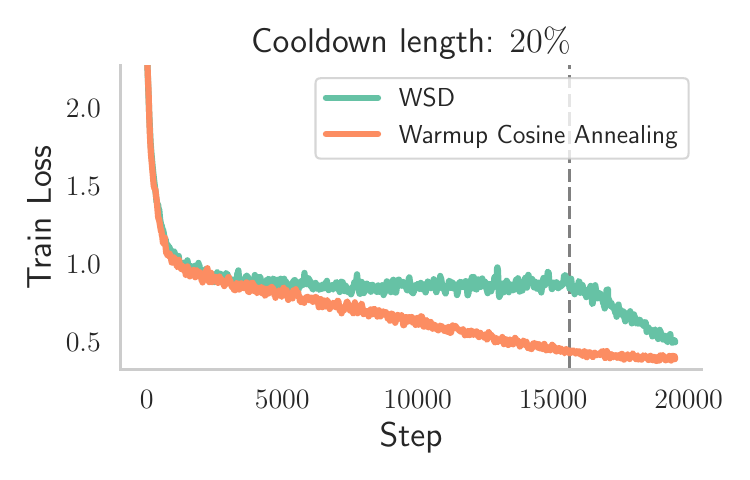}
    \includegraphics[width=0.32\textwidth]{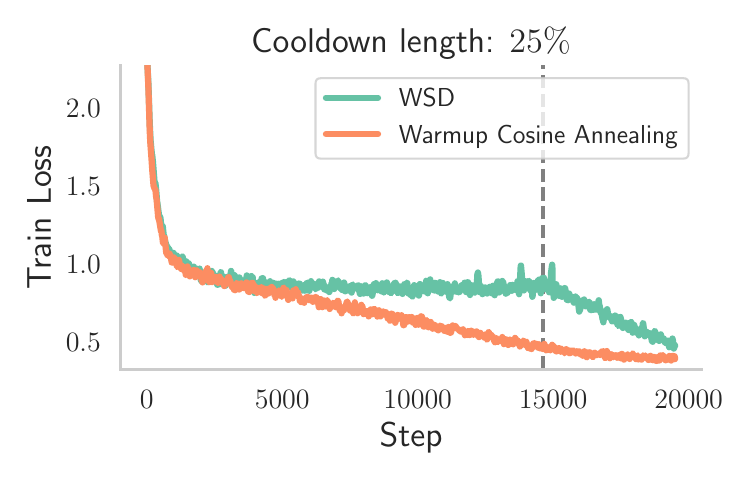}
    \includegraphics[width=0.32\textwidth]{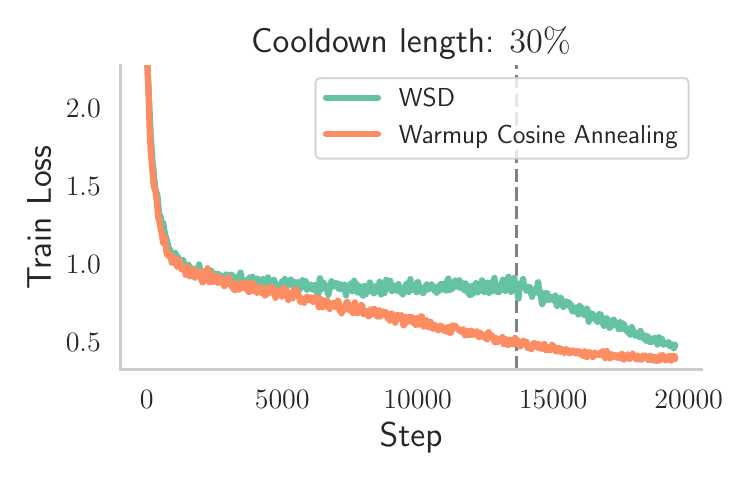}
    \includegraphics[width=0.32\textwidth]{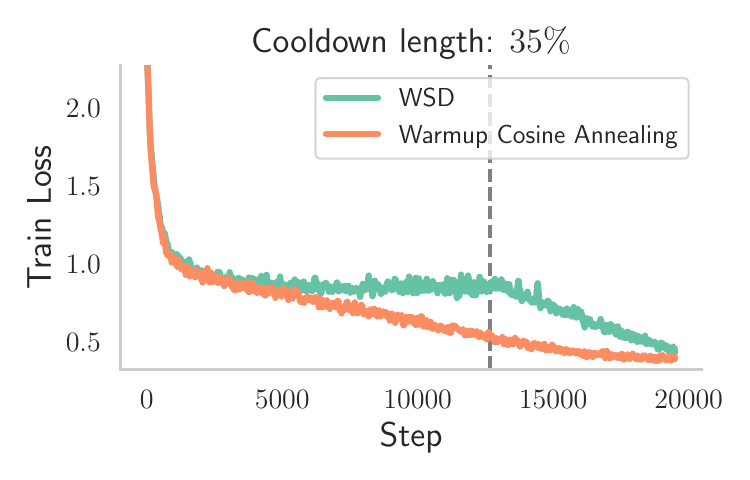}
    \includegraphics[width=0.32\textwidth]{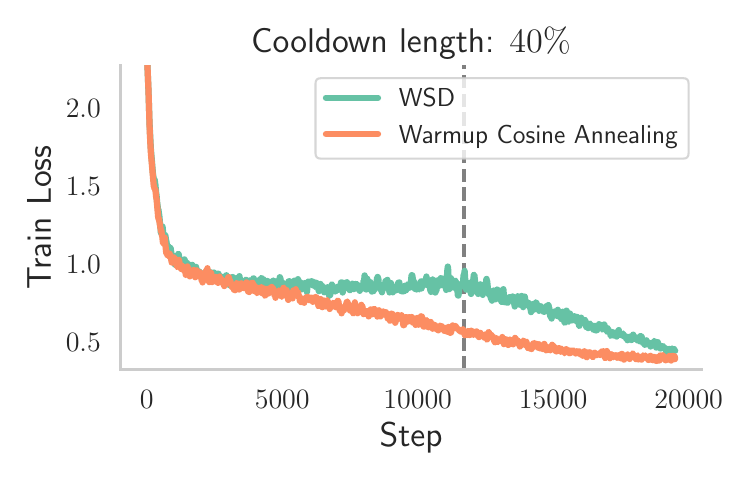}
    \caption{Visualization of the impact, on the considered CNN, of cooldown length on WSD performance, in comparison to Warmup Cosine Annealing.
    Unlike what is typically observed in transformer models, cooldown dynamics in CNNs can be slower when using the same decay length. For example, under our settings, the learning rate must decay for at least 35\% of the total training steps for the WSD performance to match that of Warmup Cosine Annealing.}
    \label{fig:Figure5}
\end{figure}

\begin{figure}[h]
    \centering
    \includegraphics[width=0.35\textwidth]{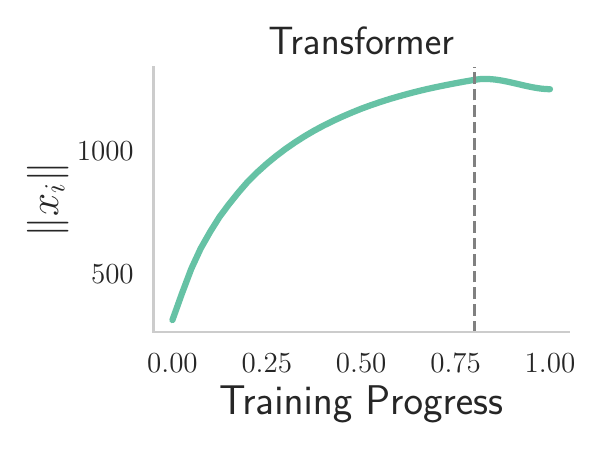}
    \hspace{0.4cm}
    \includegraphics[width=0.35\textwidth]{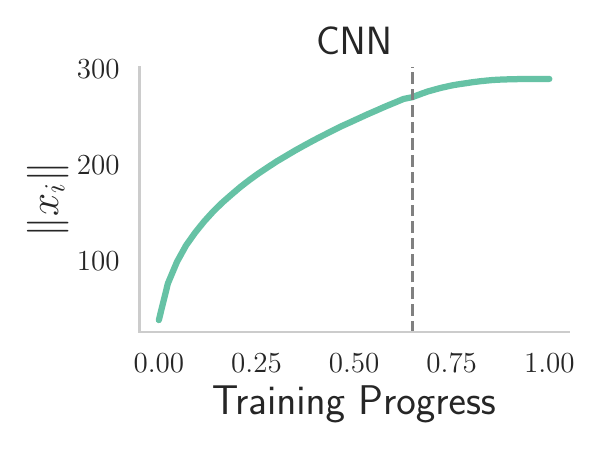}
    \caption{Parameter vector norm evolution after warmup period on an LM (left) and on a CNN (right). In both, weight updates are much larger in the stable phase (left of dashed line) than in the cooldown phase (right of dashed line).}
    \label{fig:Figure6}
\end{figure}

\begin{figure}[h]
    \centering
    \includegraphics[width=0.35\textwidth]{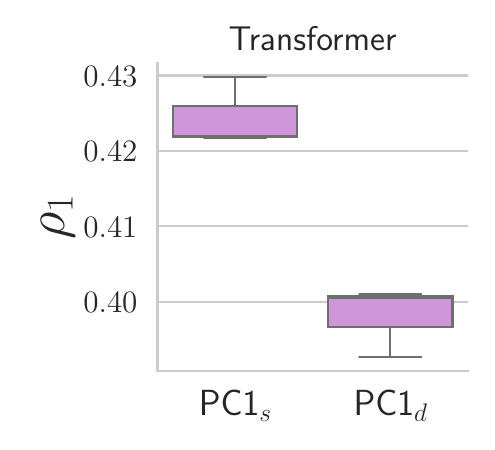}
    \hspace{0.4cm}
    \includegraphics[width=0.35\textwidth]{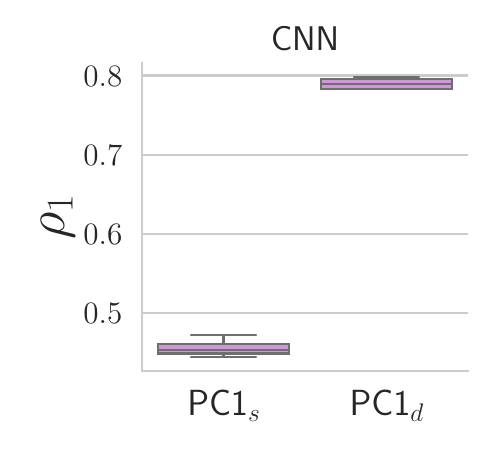}
    \caption{Box plots of the relative explained variance  \(\rho_1 = \frac{\lambda_1}{\sum_{j=1}^{n_{\text{PC}}} \lambda_j}\) of the first principal component for iterates \(x_i\) from the stable ($\text{PC1}_s$) and decay ($\text{PC1}_d$) phases, derived from multiple training runs on an LM (left) and a CNN (right). In both models and phases, the first principal component captures most of the variance, indicating that each phase is characterized by a predominant direction.}
    \label{fig:Figure7}
\end{figure}

\end{document}